\documentclass[10pt,twocolumn,letterpaper]{article}
\usepackage{cvpr}

\usepackage{soul,mathtools,amsmath,amssymb}
\usepackage[standard]{ntheorem}
\usepackage{times}
\usepackage{epsfig}
\usepackage{graphicx}
\usepackage{float}
\usepackage[algopart,noend]{algorithm2e}
\usepackage{booktabs}
\usepackage{bm}
\usepackage{graphicx}
\usepackage{csquotes}
\usepackage{siunitx}
\usepackage{nicefrac}
\usepackage{multirow}
\usepackage{float}
\DontPrintSemicolon
\makeatletter
\newcommand{\removelatexerror}{\let\@latex@error\@gobble}
\makeatother
\usepackage[pdf,dvipsnames]{xcolor}
\usepackage[inline,nomargin,final]{fixme}

\FXRegisterAuthor{aa}{aaa}{\color{orange}AA}
\FXRegisterAuthor{mf}{amf}{\color{purple}MF}

\usepackage[pagebackref,breaklinks,colorlinks,bookmarks=false]{hyperref}

\usepackage[capitalize]{cleveref}
\crefname{section}{Sec.}{Secs.}
\Crefname{section}{Section}{Sections}
\Crefname{table}{Table}{Tables}
\crefname{table}{Tab.}{Tabs.}

\newcommand{\apdA}{A}
\newcommand{\apdweights}{\mu}

\newcommand{\apds}{s}

\newcommand{\R}{\mathbb{R}}

\DeclarePairedDelimiterX{\norm}[1]{\lVert}{\rVert}{#1}
\DeclarePairedDelimiterX{\set}[1]{\lbrace}{\rbrace}{#1}

\def\cvprPaperID{*****} 
\def\confName{CVPR}
\def\confYear{2022}

\begin{document}

\title{\LaTeX\ Guidelines for Author Response}  

\title{Power-SLIC: Fast Superpixel Segmentations by Diagrams}

\author{Maximilian Fiedler\\
Zentrum Mathematik, Technische Universit\"at M\"unchen\\
D-85748 Garching bei M\"unchen, Germany\\
{\tt\small maximilian.fiedler@tum.de}
\and
Andreas Alpers\\
University of Liverpool, Department of Mathematical Sciences\\
Liverpool L69 7ZL, UK\\
{\tt\small andreas.alpers@liverpool.ac.uk}
}

\maketitle

\begin{abstract}
Superpixel algorithms grouping pixels with similar color and other low-level properties are increasingly used for pre-processing in image segmentation. In recent years, a focus has been placed on developing geometric superpixel methods that facilitate the extraction and analysis of geometric image features. Diagram-based superpixel methods are important among the geometric methods as they generate compact and sparsely representable superpixels. Introducing generalized balanced power diagrams to the field of superpixels, we propose a diagram method called Power-SLIC. Power-SLIC is the first geometric superpixel method to generate piecewise quadratic boundaries. Its speed, competitive with fast state-of-the-art methods, is unprecedented for diagram approaches. Extensive computational experiments show that Power-SLIC outperforms existing diagram approaches in boundary recall, under segmentation error, achievable segmentation accuracy, and compression quality. Moreover Power-SLIC is robust to Gaussian noise.
\end{abstract}

\section{Introduction}

Superpixels are small, non-overlapping groups of connected, perceptually homogeneous pixels. Since its introduction by Ren and Malik in 2003~\cite{renmalik}, superpixel generation has become an important pre-processing step in many imaging applications such as object localization~\cite{app1}, multi-class segmentation~\cite{app2}, optical flow~\cite{app3}, body model estimation~\cite{app4}, object tracking~\cite{app5},  depth estimation~\cite{app6}, and image denoising~\cite{app7}. 

Superpixels reduce the number of inputs for subsequent algorithms and provide meaningful image representations from which application-relevant parameters can be extracted. It is commonly desirable that superpixels adhere well to object boundaries, are  quickly generated, and compact~\cite{slicpaper, GaussianMixture, superpixelcomp}.

\vspace*{-1ex}

\begin{figure}[H]
    \centering
    \includegraphics[width=0.88\linewidth]{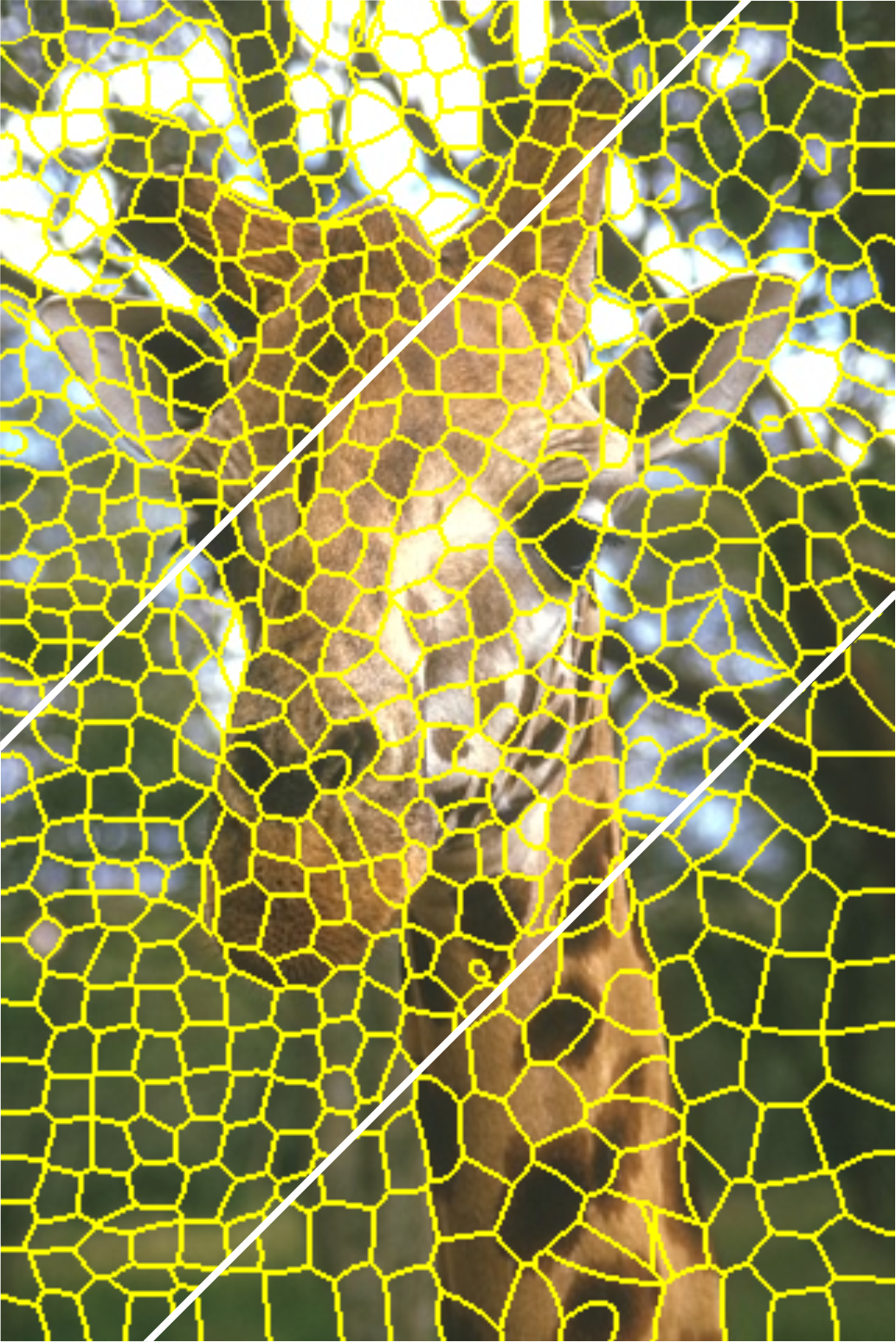}
    \caption{Superpixel segmentation using Power-SLIC with $900,$ $600,$ and $300$ superpixels.}
    \label{fig:main_image}
\end{figure}\vspace*{-2.5ex}

Traditionally, superpixel algorithms are pixel-based because they operate on the pixel level, generating regions delimited by sets of pixels. Such pixel-level superpixel representations can often approximate small details and adhere well to boundaries. However, they are generally non-compact and free of higher-level geometric information.

Addressing this aspect, recent studies~\cite{SLICpolygon, paperconvexpolygons, kippi, eccpd,kurlin2020}, have focused on developing geometric superpixel methods. These methods  partition the image into small, preferably size-controllable, geometric objects, e.g., polygonal cells. As these objects `live' in the continuous world, they provide a resolution-independent superpixel representation and facilitate geometric manipulations, feature extractions, and other higher-level geometric post-processing tasks. The applicative potential of geometric superpixel methods has been demonstrated for diverse computer vision tasks such as object contouring~\cite{kippi, eccpd}, image vectorization~\cite{piecewisepoly}, image compression~\cite{kurlincompression, eccpd}, and segmentations of scenes containing man-made environments or other strong geometric signatures~\cite{paperconvexpolygons}.

Many pixel methods, e.g.,~\cite{slicpaper,SLICmanifold, IntrinsicManifoldSLIC, vcells}, employ Voronoi diagrams as intermediate structure, but they depart from it to better control the sizes and shapes of the superpixels.  In constrast, the geometric methods Varane~\cite{paperconvexpolygons, Duan} and, most recently, ECCPD~\cite{eccpd} return diagram cells as superpixels that are compact and sparsely representable. However, both of these diagram approaches employ time-consuming post-optimization to control the superpixel sizes and shapes. Motivated by unifying and extending the diagram approaches, we propose a novel method, called Power-SLIC, which handles non-linear boundaries and brings the speed of state-of-the-art pixel methods to the geometric world.

\subsection{Technical Contribution}
The main technical contribution of our work is that we bridge the gap between pixel and geometric superpixel methods by utilizing the following three techniques: (i) a deep connection between constrained clustering and diagrams from~\cite{briedengritzmann10} (see \cref{thm:wblsa:correspondence}) to control the diagram cell sizes,  (ii) the heuristic~\cite{APDheuristic} based on this, and (iii) SLIC's initialization \& assignment step.
 
\textbf{Contribution to the geometric world:} A diagram method that, for the first time, has a  speed similar to fast pixel-based methods and that outperforms existing approaches in  boundary recall, under segmentation error, achievable segmentation accuracy, and compression quality. To our knowledge, this is the first geometric approach capable of generating non-linear superpixel boundaries. 

\textbf{Contribution to the pixel world:} A fast and noise-robust method that clusters pixels into superpixels but provides the benefits of diagram approaches such as providing sparse geometric superpixel representations by diagrams, facilitating  geometric image operations, and providing relatively compact and regular superpixels at the same time.

\subsection{Related Works}
We briefly review superpixel algorithms classified into two groups: pixel methods and geometric methods. For a comprehensive evaluation see~\cite{superpixelcomp, benchmark}.

\mfnote{Ist Noise-Robust hier wirklich der richtige Fokus? }

\paragraph{Pixel methods}
Many popular state-of-the-art superpixel generation methods are pixel-based. A large number of pixel methods, such as SLIC~\cite{slicpaper}, VCells~\cite{vcells}, TurboPixel~\cite{turbopixel}, and  SEEDS~\cite{seeds}, employ clustering algorithms refining the clusterings until a convergence criterion is satisfied. SLIC, and variants such as SLIC0~\cite{sliczero},  SNIC~\cite{SLICpolygon},  Manifold-SLIC~\cite{SLICmanifold}, and Intrinsic Manifold-SLIC~\cite{IntrinsicManifoldSLIC}, are particularly popular due to their simplicity and high performance. 

Other algorithms, such as  ERS~\cite{ers} and Normalized Cuts~\cite{renmalik}, formulate the superpixel generation task as an optimization problem on a graph structure.  

It has been noted several times (e.g., in~\cite{superpixelnoise, noisepaper2, superpixelcomp}) that the  performance of superpixel algorithms degrades severely as  noise levels increase. To address this issue,~\cite{noiserobustSLIC} proposes a Mahalanobis-based $k$-means variant (NR-SLIC). The Mahalanobis distance is also used in \cite{GaussianMixture} and \cite{anisotropicsuperpixel}, which introduce a Gaussian mixture model (GMMSP) and a unimodular Gaussian generative model, respectively. Gaussian mixture models based on Bayesian or Normal-Inverse Wishart priors for the covariances are given in~\cite{bayesiansuperpixel1, bayesiansuperpixel2}. Common to these approaches is that they consider Mahalanobis distances for both the pixels' spatial and color components. Their clusterings in 5D can be viewed as resulting from five-dimensional generalized balanced power diagrams (GBPDs)~\cite{philmag}. As projections from 5D into 2D, their spatial cell boundaries are, however, rather irregular; see GMMSP and NR-SLIC in~\cref{fig:algorithm_comparison_and_noise}. Our approach utilizes the Mahalanobis distance only for the spatial components of the pixels, resulting, as we will demonstrate, in two-dimensional GBPDs with smooth, quadratic cell boundaries.

\paragraph{Geometric methods}
Geometric methods (also known as resolution-independent or polygonal decomposition methods) aim at extracting continuous parameter representations of the superpixel boundaries. 

The algorithm from~\cite{paperconvexpolygons}, referred to as Varane~\cite{Duan}, partitions the image into convex polygonal superpixels by building Voronoi diagrams that conform to preliminarily detected line segments.  The color values inside these Voronoi cells can be approximated; for example, \cite{piecewisepoly} proposes an approximation by polynomials. More general than Voronoi diagram cells, ECCPD from~\cite{eccpd} proposes to segment images into power diagram cells. The algorithm iterates between updates of sites and weights and applies an edge-alignment optimization step. 

Meshes of convex polygonal cells that are guaranteed to have no small angles are generated in~\cite{kurlincompression}. An improved method is given in~\cite{kurlin2020}. The algorithms SNIC and SNICPOLY from~\cite{SLICpolygon} produce superpixels with piecewise linear boundaries. KIPPI from~\cite{kippi} uses a kinetic approach that extends line segments until they meet each other. KIPPI can generate polygons of different sizes, avoiding oversegmentations of large uniform areas, but it does not allow  explicit control over the number of superpixels.  A Delaunay Point Process~(DPP), which segments images into triangular patches, is proposed in~\cite{Delaunay}. 

Noticeably, many algorithms compute an intermediate diagram structure. Diagrams are returned as final superpixel segmentation  only in Varane and ECCPD.  Therfore, we refer to them as diagram approaches. They return Voronoi and power diagrams, respectively, which have linear boundaries. In contrast, our diagram class of GBPDs contains the latter diagram classes as subclasses. Boundaries can be quadratic, but the cell convexity may be lost.

\section{Overview of our Approach}
\begin{figure}[H]
    \vspace*{-2ex}
    \centering
    \includegraphics[width=1\linewidth]{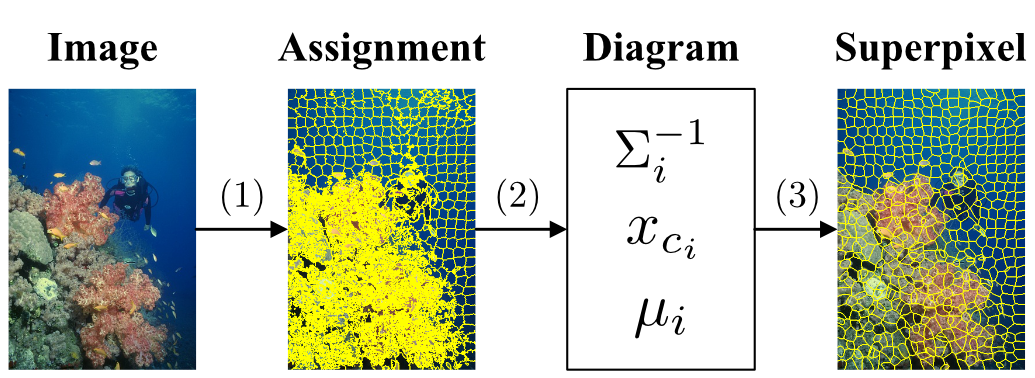}
    \caption{Main steps of Power-SLIC: (1) Perform SLIC's initialization \& assignment step. (2) Extract diagram parameters from the assignment yielding a geometric representation of the superpixels. (3) Assign each pixel to its containing diagram cell (yielding a pixel representation of the superpixels). }
    \label{fig:power-slic-flow-chart}
\end{figure}

We borrow the initialization \& assignment step from SLIC (Simple Linear Iterative Clustering)~\cite{slicpaper} to obtain possibly non-connected groups of pixels, which typically have a high boundary adherence but fuzzy boundaries. In contrast to SLIC's post-processing, which guarantees connected superpixels, we compute a generalized balanced power diagram (GBPD) based on the spatial Mahalanobis distance determined for every previously obtained group of pixels. The cells of the GBPD will ultimately be the resulting superpixels. As the computation of GBPDs can be time-consuming, we approximate it to achieve a competitive running time. A high-level visualization of Power-SLIC is illustrated in \cref{fig:power-slic-flow-chart}.

\vspace*{-2ex}

SLIC's assignment phase ideally suits our purposes, as it constructs a five-dimensional Voronoi diagram, a special case of a GBPD, with linear boundaries. Our approach aims at approximating and regularizing this 5D structure by a GBPD with quadratic cell boundaries in the two spatial dimensions.

\section{SLIC's Initialization \& Assignment Phase}
SLIC~\cite{slicpaper} is seen as one of the most popular and widely used methods for superpixel generation. Given a color image of~$N$ pixels, SLIC clusters the pixels within the five-dimensional space $(x_p,l_p)$, with $x_p$ denoting the spatial position of pixel~$p$ (a two-dimensional vector) and  $l_p$ its color (a three-dimensional vector) in the CIELAB color space. The algorithm is divided into an initialization \& assignment and a post-processing phase. The latter phase ensures  connectivity by reassigning near-by superpixels. As we do not make use of this phase, we omit a description.  

\paragraph{Initialization \& Assignment} SLIC takes as its input two arguments~$k$ and~$m$, with~$k$ specifying the targeted number of superpixels and~$m$ denoting a compactness parameter. Throughout this paper, we use $m=10$ as this is SLIC's default value. Initially, SLIC samples cluster centers with spatial positions from a square grid spaced $h = \sqrt{N/k}$ pixels apart. Subsequently,
%
 pixels are assigned to their closest cluster center. However only local assignments of pixels lying within a $2h \times 2h$ spatial window around a cluster center are considered to improve performance. Proximity is measured by a weighted Euclidean distance. The second parameter, $m$, determines the weight of the Euclidean distance in the color space. For a pixel $p$ and a center~$c$, the squared distance~$d^2$ is given by:
\begin{align}
    d^2(p,c) =\norm{x_p-x_c}_2^2 
    + \frac{h^2}{m^2} \cdot\norm{l_p-l_c}_2^2.
\end{align}
 As in traditional $k$-means, the centers are updated as the centroids of the updated clusters, and the process is iterated until some stopping criterion is fulfilled. 

\section{Generalized Balanced Power Diagrams}
Generalized balanced power diagrams (GBPDs)~\cite{philmag} generalize the concept of power diagrams, which themselves generalize the famous Voronoi diagrams. Here, we are given a set of $k$ distinct sites $S = \set{ \apds_1, \ldots, \apds_k} \subseteq \R^d$. Each site is equipped with a local ellipsoidal norm defined by a positive definite symmetric matrix $\apdA_i \in \R^{d \times d}$.  The norm  $\norm{\cdot}_{A_i}$ is defined as  $\norm{ x }_{A_i} := \sqrt{ x^\top A_i x }$ for $x \in \R^d,$ $i\in[k]$. Moreover, let  $\mu_1,\dots\mu_k$ be scalars serving as additional balancing parameters.
 
 For every $i \in [k]$, the generalized balanced power diagram cell is defined as:
 \begin{align}
 \begin{split}
 	P_i := \{ x \in \R^d:\:  &\norm{x - \apds_i}_{\apdA_i}^2 + \mu_i \\ &\leq   \norm{x - \apds_l}_{\apdA_l}^2 + \mu_l, ~~\forall l \in [k]\}.
 	\end{split}\label{eq:GBPDcell}
 \end{align}
We  call the $k$-tuple of cells $(P_1, \ldots, P_k)$ a generalized balanced power diagram (GBPD). Throughout this paper, we consider only two-dimensional GBPDs, although our approach canonically extends to images and GBPDs in arbitrary dimensions.
 
Further, we remark that we obtain power diagrams in the special case in which each $A_1,...,A_k$ are the identity matrix.  If additionally $\mu_1=\cdots=\mu_k$, we obtain Voronoi diagrams.

\section{Optimal Power-SLIC and Power-SLIC}

Given the assignment~$L$ computed by SLIC's initialization \& assignment phase, we propose computing the following statistics for each label $i$ in $L$: the spatial center~$x_{c_i},$ the spatial covariance matrix~$\Sigma_i \in \R^{2 \times 2},$ and the area~$\kappa_i$ of each possibly non-connected component $C_i=\{x_p: L(p)=i\}.$ 
From this we compute a GBPD with sites~$s_i=x_{c_i}$ and matrices $A_i=\Sigma^{-1}_i$, such that the area of its cells~$P_i$ equals or approximates~$\kappa_i,$ $i\in[k].$ The cells~$P_1,\dots,P_k$ of the GBPD will be the resulting superpixels.

\paragraph{Optimal Power-SLIC} We first present an algorithm to compute GBPDs as described above. As in~\cite{philmag}, we can achieve this by exploiting a strong relation between  constrained clustering and (anisotropic) power diagrams shown in~\cite{briedengritzmann10} (see also~\cite{briedengritzmann12}).  We consider the following linear program:
\begin{equation}
	\tag{P}
	\begin{alignedat}{3}
		&\min&& \sum_{i=1}^{k} \sum_{j=1}^{N} \xi_{ij}\norm{x_{p_j}-s_{i}}_{A_{i}}^{2}  \\
		&\,\text{s.t.}&&\begin{aligned}[t]
			\sum_{i=1}^{k} \xi_{ij} &= 1, &\quad j \in [N],&\\
			\sum_{j=1}^{n} \xi_{ij} &= \kappa_{i}, &\quad i \in [k],& \\
			\xi_{ij} & \ge 0,   &\quad i \in [k],\ j \in [N]&.
		\end{aligned}
	\end{alignedat}
	\label{lp:primal}
\end{equation}
Its dual is given by:
\begin{equation}
	\tag{D}
	\begin{alignedat}{3}
		&\max&&  \sum_{j=1}^{N}  \eta_{j} - \sum_{i=1}^{k} \kappa_{i}\mu_{i}  \\
		&\,\text{s.t.}&&\begin{aligned}[t]
			\eta_{j} & \le \norm{x_{p_j}-s_{i}}_{A_{i}}^{2} + \mu_{i} ,   &\quad i \in [k],\ j \in [N].&
		\end{aligned}
	\end{alignedat}
	\label{lp:dual}
\end{equation}

The problem modeled by~\labelcref{lp:primal} is the balanced least-squares assignment problem. The task is to assign $x_{p_j}$ to sites in such a way that precisely~$\kappa_i$ of them are assigned to site~$s_i$, for each $i\in[k],$ while the sum of squared distances of the assigned $x_{p_j}$ to their assigned sites, measured via the norm $||\cdot||_{A_i}$, is minimized. As the constraint matrix of the problem is totally unimodular, there will be a binary optimal solution to~\labelcref{lp:primal}, i.e., in this solution, $\xi^*_{ij}$ is either 0 or 1 where $\xi^*_{ij}=1$ if, and only if, $x_{p_j}$ is assigned to site~$s_i.$ 

Optimal solutions to~\labelcref{lp:primal} can be characterized in terms of GBPDs. To state this precisely, we apply the following notion: a diagram is said to induce a binary assignment if, and only if, all $x_{p_j}$ assigned to the same cluster lie in the same diagram cell.

\begin{theorem}[Special case of \cite{Brieden2017}] \label{thm:wblsa:correspondence}
		A balanced assignment is a basic solution of the linear program \labelcref{lp:primal} if and only if there exists a parameter vector $\apdweights \in \R^k$ such that
	the GBPD with parameters $(A_i, s_i, \mu_i)$, $i \in [k]$, induces this binary assignment. Moreover, with a solution $(\mu^*,\eta^*)$ of \cref{lp:dual} that fulfills strict complementary slackness, we can choose $\mu=\mu^*.$ 
\end{theorem}

Hence, when optimally assigning pixels to superpixel centers, i.e., for $s_i=x_{c_i},$ superpixel areas prescribed to equal~$\kappa_i,$ and distance matrices $A_i = \Sigma_i^{-1},$ \mfnote{fix} the resulting superpixels can be characterized by cells of a GBPD. Moreover, the diagram parameters are extracted from the dual solution and yield a low-dimensional, resolution-independent representation of the superpixels (an algebraic representation of the boundary between two neighboring cells is provided by the quadratic equation that results  from changing the respective inequality relation in~\cref{eq:GBPDcell} to an equality relation).

In terms of computational complexity, we remark that~\labelcref{lp:primal} has $Nk$ variables and $2N+k$ constraints. The linear program can, however, be solved in $\mathcal{O}(N^3)$ using the Hungarian method, for example. In our evaluation, we use the dual simplex of Gurobi~\cite{gurobi}. To improve performance, we apply a locality assumption similar to SLIC by adding only those variables $\xi_{ij}$ for which~$p_j$ lies within a regional window of the center~$c_i$. To ensure that every pixel is assigned, we choose this window to be the smallest bounding box of~$C_i$. The resulting algorithm, called Optimal Power-SLIC, is summarized in Algorithm~\labelcref{alg:optimal_power_slic}. 

We found that building the model takes up most of the time, while solving the model usually requires less than $10\,$s per image. This suggests that more significant speed-ups can be achieved by avoiding Gurobi calls, e.g., by using a more specialized and adapted solver, such as min-cost flow.

\begin{figure}[t]
   \renewcommand\figurename{Algorithm}
\removelatexerror
\begin{algorithm}[H]
 \KwData{Image, targeted number~$k$ of superpixels.}
 \KwResult{Assignment, GBPD.}
  Run SLIC's initialization \& assignment phase with parameters $k$ and $m=10$ to obtain assignment $L$.\;
 \For{each pixel label $i$ in $L$}{
    Compute spatial center~$x_{c_i}$, covariance matrix~$\Sigma_i,$ and area $\kappa_i$ of~$C_i$.\;
    \For{each pixel $p_j$ inside of $C_i$'s smallest bounding box}{
        Generate the variable $\xi_{ij}$ of \labelcref{lp:primal}.\;
    }
 }
 Solve~\labelcref{lp:primal} for $s_i=x_{c_i}$ to retrieve both primal~$\xi_{ij}^*,$ and dual solution $\mu_i^*,\eta_j^*,$  $i\in[k],$ $j\in[N]$.\;
 Set $\mu = (\mu^*_1,\dots,\mu^*_k)$.\;
 \Return{assignment (computed from the $\xi^*_{ij}$) and diagram parameters $(\Sigma^{-1}_i,x_{c_i},\mu_i)$, $i \in [k]$.} 
\end{algorithm}
 \caption{Optimal Power-SLIC.}
  \label{alg:optimal_power_slic}
\end{figure}\vspace*{-2ex}

\paragraph{Power-SLIC} To match the SLIC linear running time, we now introduce a heuristic version of Optimal Power-SLIC, which we call Power-SLIC. The main ideas are rooted in two key observations. 

First, we observe that most of the computation time in Optimal Power-SLIC is spent on computing  $\mu_1,\dots,\mu_k$ while the sites and norm matrices of the diagram are essentially readily available. To effect a significant speed-up, we replace the computation of the optimal~$\mu_i$ with an estimation. The idea is to think of $\mu_i$ as a scaling factor and estimate it such that the ellipsoid given by the inverse covariance matrix $A_i = \Sigma^{-1}$ has an area equal to $\kappa_i$. Thus, we want to choose $\mu_i$ such that $\mathrm{Area}(\norm{x}_{\Sigma_i^{-1}}^2 + \mu_i \le 0) = \kappa_i$ which is equivalent to $\mathrm{Area}(\norm{x}_{\Sigma_i^{-1}}^2  \le -\mu_i) = \kappa_i$.

Since the area of such an ellipsoid is given by $\pi \sqrt{\mu_{i}^{2}\det{\Sigma_{i}}}$, we solve $	\kappa_{i} = \pi \sqrt{\mu_{i}^{2}\det{\Sigma_{i}}}$ for $\mu_i$ and choose the appropriate solution for which $-\mu_i > 0$, hence
\begin{equation} \label{eq:adapting_sigma_heuristic}
    -\mu_i = \frac{\kappa_{i}}{ \pi \sqrt{\det(\Sigma_i)}}.
\end{equation}
Our evaluation will show that this choice provides a very good compromise between speed and boundary adherence. (For a use of ~\cref{eq:adapting_sigma_heuristic} in a materials science context, see~\cite{APDheuristic}.)

The second observation relates to the fact that the superpixel statistics can be extracted efficiently from the final SLIC assignment iteration without requiring any additional passes over the image data. While this is also true for Optimal Power-SLIC, it becomes particularly relevant with fast implementations of Power-SLIC. In fact, the spatial centers~$x_{c_1},\dots,x_{c_k}$ can be obtained after the final SLIC iteration simply by projecting $c_1,\dots,c_k$ into the spatial image space by dropping their color components. The spatial covariance matrices $\Sigma_1,\dots,\Sigma_k,$ can be computed together with the centers within the final iteration of the SLIC assignment, using a single pass algorithm for the variance.

With these parameters, Power-SLIC computes a final assignment that aims at assigning each pixel~$p_j$ to its closest spatial center~$x_{c_i},$ $i\in[k].$ Distances are measured in terms of  $||x_{p_j}-x_{c_i}||_{\Sigma_i^{-1}}^2 + \mu_i$.  As in SLIC, we consider only those pixels lying in a $2h \times 2h$ window around the cluster center for the final assignment. In a cleaning step, the still unassigned pixels are assigned to its containing diagram cell. In our experiments, this cleaning step was hardly ever required. Thus, not even a naive implementation would affect the practical running time. A summary of Power-SLIC is given in Algorithm~\labelcref{alg:power_slic}. 

\begin{theorem}
Power-SLIC runs in $\mathcal{O}(N)$.
\end{theorem}

\begin{proof} SLIC's initialization \& assignment phase runs in~$\mathcal{O}(N)$, the computation of the diagram parameters is on the fly, and the diagram cell computation for obtaining the superpixels is essentially one additional assignment iteration with a modified distance measure. Thus, we obtain an overall running time of $\mathcal{O}(N),$ which is independent of the number of superpixels~$k$. 
\end{proof}

\begin{figure}[t]
   \renewcommand\figurename{Algorithm}
\removelatexerror
\begin{algorithm}[H]
\KwData{Image, targeted number~$k$ of  superpixels.}
 \KwResult{Assignment, GBPD.}
 Run SLIC's initialization \& assignment phase with parameters $k$ and $m=10$ to obtain spatial centers $x_{c_i}$ and spatial covariance matrices~$\Sigma_i,$ $i\in[k].$\;
 Compute $\mu_i$ using \cref{eq:adapting_sigma_heuristic}.\;
 Set $L(p_j)= -1$ and $D(p_j) = \infty$ for each pixel $p_j$.\;
\For{each cluster $i \in [k] $}
    {
        \For{each pixel $p_j$ in a spatial $2h \times 2h$ window around $x_{c_i}$}
        {   
            \If{$ \norm{x_{p_j}-x_{c_i}}^2_{\Sigma_i^{-1}} +\mu_i  < D(p_j) $ }{
            Set $D(p_j) = \norm{x_{p_j}-x_{c_i}}^2_{\Sigma_i^{-1}} +\mu_i $.\;
            Set $L(p_j) = i$.\;
            }
        }
    }
    \For{each pixel $p_j$ with $L(p_j) = -1$}
    {   
        Determine cell $P_i$ of the GBPD containing $p_j$ and set $L(p_j) = i$.\;
    }
 \Return{L and diagram parameters $(\Sigma^{-1}_i,x_{c_i},\mu_i)$, $i \in [k]$.}
\end{algorithm}
 \caption{Power-SLIC.}
  \label{alg:power_slic}
\end{figure}\vspace*{-2ex}

\section{Evaluation}

\begin{figure*}
\centering
\includegraphics[width=\textwidth]{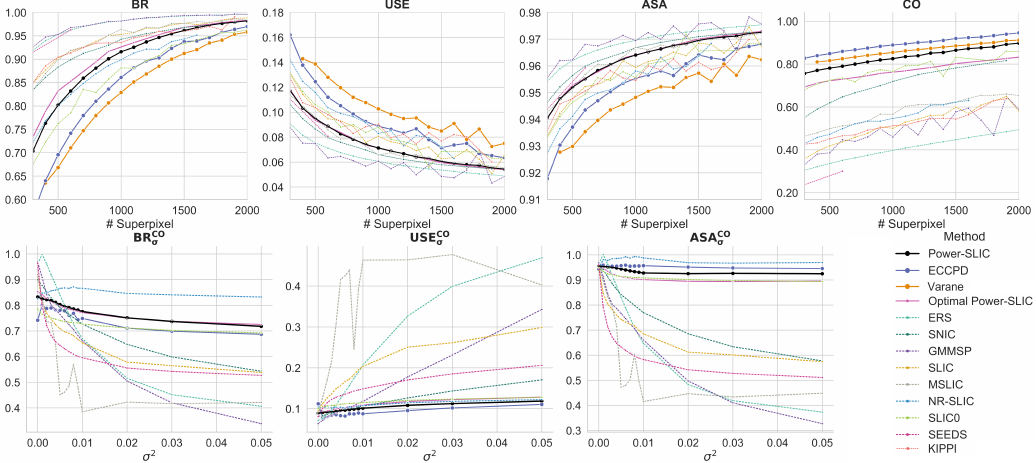}
\caption{Top: Quantitative comparison of multiple superpixel algorithms. Bottom: Noise robustness of multiple superpixel algorithms  and $k=600$. Here, no results are shown for Varane and KIPPI because the superpixels could not be obtained for noisy images.} 
\label{fig:algorithm_comparison_and_noise}
\end{figure*}

In this section, we compare Power-SLIC and Optimal Power-SLIC on the Berkeley Segmentation Data Set 500 (BSDS500)~\cite{bsds500} with the state-of-the-art superpixel algorithms SLIC~\cite{slicpaper}, SLIC0~\cite{sliczero}, NR-SLIC~\cite{noiserobustSLIC}), GMMSP~\cite{GaussianMixture}), Varane~\cite{paperconvexpolygons}, ECCPD~\cite{eccpd}, KIPPI~\cite{kippi}, ERS~\cite{ers}, SNIC~\cite{SLICpolygon}, and SEEDS\cite{seeds}.

\paragraph{Quantitive Evaluation}
For our quantitive evaluation, we compute boundary recall, undersegmentation error, achievable segmentation accuracy, and compactness for~$k$ number of superpixels between 300 and 2,000. Additionally, we report the running times and  peak signal-to-noise ratios for image compression for various~$k$. Finally, we measure the noise robustness by considering the trade-off between boundary recall and compactness. 

Boundary recall~($BR$) measures the percentage of detected ground truth boundary pixels. With $T_p$ and $F_p$ denoting the number of true positives and false negatives, respectively, $BR$ is defined as $BR = \nicefrac{T_p}{(T_p + F_n)}.$ In practice, we follow~\cite{slicpaper} and classify a ground truth boundary pixel as a true positive if its spatial two-neighborhood contains a superpixel boundary pixel. Accordingly,~$F_p$ equals the number of superpixel boundary pixels not in a spatial two-neighborhood of a ground truth boundary pixel.

We also compute the undersegmentation error (USE), which increases when different ground truth regions are grouped into the same superpixel. With $\mathcal{G}$ denoting the set of segments in the ground truth and $\mathcal{S}$ denoting the set of superpixels, USE can be defined as:
\begin{equation}
    USE = \frac{1}{\sum_{G \in \mathcal{G}} |G|} \left( \sum_{S \in \mathcal{S}} \sum_{G \in \mathcal{G}} \min(|S \cap G|,|S \setminus G|)\right). 
\end{equation}

The achievable segmentation accuracy (ASA) provides an upper bound on the best obtainable segmentation accuracy of the image when segmenting on the superpixel level. It is defined as 
\begin{equation}
    ASA = \frac{\sum_{S \in \mathcal{S}} \max_{G \in \mathcal{G}} |S \cap G| }{\sum_{G \in \mathcal{G}} |G|}.
\end{equation}

To evaluate compactness, we compute the compactness of the superpixels as introduced in~\cite{compactnessmetric}. Let $b_i$ and $\kappa_i$ denote the $i$th superpixel's total number of boundary pixels and area. With $Q_i$ the iso-perimetric quotient, compactness $(CO)$ is defined as:
\begin{equation}
    Q_i = \frac{4 \pi \kappa_i}{b_i^2}, \qquad   CO = \sum_{i=1}^k Q_i \frac{\kappa_i}{N}.
\end{equation}
Compactness essentially measures the similarity of the superpixels to a disc. 

\Cref{fig:algorithm_comparison_and_noise} provides a comparison among the superpixels algorithms. First, note that the algorithms Varane, ECCPD, and (Optimal) Power-SLIC are the only algorithms that compute a low-dimensional diagram representation of the superpixels. As these superpixels are diagram cells, they result in high compactness values.

While Power-SLIC generates slightly less compact superpixels than Varane and ECCPD, it outperforms the  diagram approaches in BR, USE, and ASA for any number~$k$ of superpixels. Remarkably, Power-SLIC's performance in these categories falls into the range achieved by pixel methods.

\begin{figure*}
    \centering
    \includegraphics[width=\textwidth]{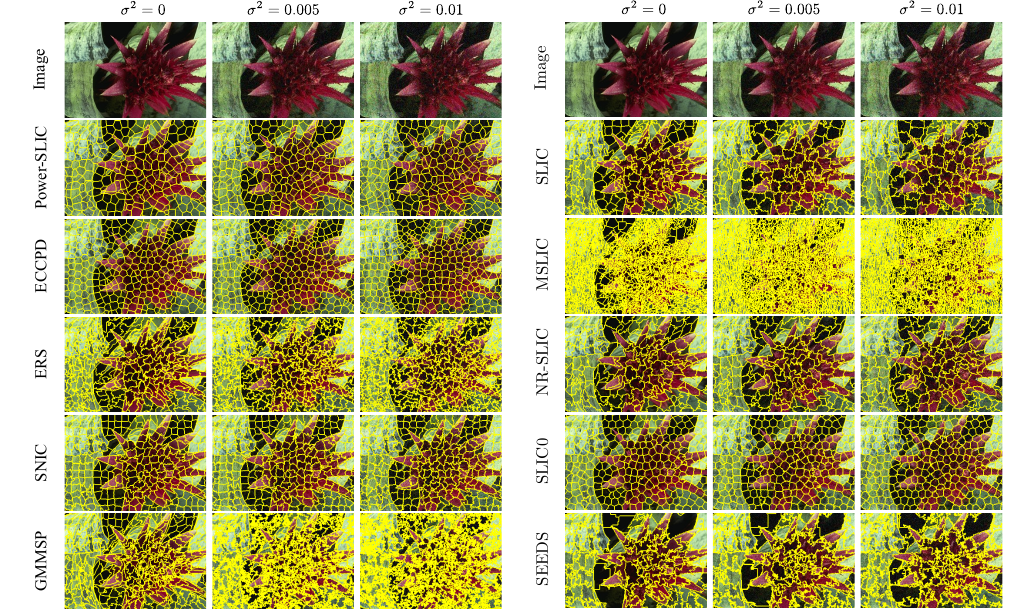}
    \caption{Visual comparison of five  superpixel algorithms for various levels of noise ($k=700$).}
    \label{fig:algorithm_visual_comparison}
\end{figure*}

Pixel methods are often highly sensitive to changing image conditions and can fail when small amounts of noise are present in the image. For example, looking at GMMSP in \cref{fig:algorithm_visual_comparison} its superpixel segmentation degenerates quiet considerably if noise is added: The algorithm identifies too many image boundaries, and almost all pixels are identified as boundaries. While this results in a high boundary recall by definition, the segmentation deteriorates and the compactness value decreases significantly. It is therefore reasonable to penalize the measures BR, ASA, USE for a decreasing compactness value; see~\cite{eccpd}, which also relates the measures to the compactness value. Let $CO_\sigma$ denote the compactness value for the noise level $\sigma^2$. Similarly, we denote by $BR_\sigma$, $ASA_\sigma$, $USE_\sigma$ the measures at noise level~$\sigma^2$. We define 
\begin{equation*}
    \begin{gathered}
    BR^{CO}_\sigma = BR_\sigma \cdot  \frac{ CO_\sigma}{CO_0}, \quad
    ASA^{CO}_\sigma = ASA_\sigma \cdot \frac{ CO_\sigma}{CO_0}, \\
    USE^{CO}_\sigma = USE_\sigma \cdot \frac{ CO_0}{CO_\sigma}. 
\end{gathered}
\end{equation*}
In other words, we consider the compactness relatively to the compactness for noise-free images so that the measures do not depend on the compactness of the superpixel segmentation for noise-free images. For a perfectly noise-robust superpixel algorithm, we expect the measures to be constant and therefore independent of~$\sigma$. \Cref{fig:algorithm_comparison_and_noise} summarizes the results for $k=600$ (a similar behavior is observed for other typical values of $k$). We observe that NR-SLIC, an algorithm specifically designed for noisy images, performs best overall. On the other hand, those algorithms with very high boundary recall on clean images tend to deteriorate quickly, even for small amounts of noise ($\sigma^2 < 0.01$), mainly because they detect too many \enquote{image boundaries} in the noisy cases (see also \cref{fig:algorithm_visual_comparison}). Power-SLIC and ECCPD are relatively robust to noise, and their performance decreases only slightly with increasing noise.

We now focus on two aspects of the diagram approaches. First, we measure computational speed,  a bottleneck of diagram methods  so far. Second, we assess the compression quality achieved by the highly sparsely-encoded superpixel segmentation provided by diagrams. 

Timings are shown in \cref{tb:running_times_psnr}. Power-SLIC is several orders of magnitude faster than the other diagram approaches and its running time increases only slightly with~$k.$

\begin{table}
\small
\centering
\resizebox{0.9\linewidth}{!}{%
\begin{tabular}{@{}llrrrr@{}}
\toprule
                                     &            & $k$=500         & $k$=1,000       & $k$=1,500       & $k$=2,000       \\ \midrule
\multirow{3}{*}{\begin{tabular}[c]{@{}l@{}}Time\\ (in s)\end{tabular}} & Varane     & 2.549           & 10.191          & 18.620          & 47.022          \\
                                     & ECCPD      & 28.703          & 96.408          & 149.929         & 171.321         \\
                                     & Power-SLIC & \textbf{0.070}  & \textbf{0.074}  & \textbf{0.078}  & \textbf{0.080}  \\ \midrule
\multirow{3}{*}{PSNR}                & Varane     & 29.667          & 29.998          & 30.271          & 30.318          \\
                                     & ECCPD      & 29.716          & 30.101          & 30.369          & 30.259          \\
                                     & Power-SLIC & \textbf{30.201} & \textbf{30.578} & \textbf{30.840} & \textbf{31.027} \\ \bottomrule
\end{tabular}%
}
\caption{Top: Running times on a laptop with an Intel Core i7 5600U processor. Bottom: Peak signal-to-noise ratios.}
\label{tb:running_times_psnr}
\end{table}

As mentioned above, diagram approaches are particularly suited to provide sparse image encodings on which one can base subsequent higher-level computer vision tasks or employ them for image compression~\cite{eccpd}. Using Power-SLIC, for instance, one can represent and store an image by specifying~$9k$ (floating point) parameters: for each superpixel, two for its cell site, three for covariance matrix, one for the size parameter, and three for the average color.

Following papers such as~\cite{eccpd}, we evaluate the quality of the generated sparse image representation by reporting  the peak signal-to-noise ratio (PSNR) between the original and compressed image. With $x_{p_j}$ and $\hat{x}_{p_j}$  denoting the color value of the $j$-th pixel in the original and compressed image, respectively. the PSNR is defined as
\begin{align}
    MSE &= \frac{1}{3N} \sum_{j =1}^N \norm{x_{p_j} - \hat{x}_{p_j}}_2^2, \\
    PSNR &= 10 \cdot \log_{10}\left( \frac{255^2}{MSE}\right).
\end{align}
Larger PSNR values indicate a better quality of the compressed image. The resulting PSNR values for the BSDS500 dataset are shown in \cref{tb:running_times_psnr}. In all cases, Power-SLIC gives the highest PSNR values despite having the smallest running times. It seems that the anisotropic diagram structure, involving quadratic boundaries, allows Power-SLIC to improve on Varane and ECCPD, both of which use less flexible diagram classes.

\begin{table}
\centering
\resizebox{0.9\linewidth}{!}{%
\begin{tabular}{@{}lcccc@{}}
\toprule
           & Diagram & Fast & \multicolumn{1}{l}{Compact} & \multicolumn{1}{l}{Noise-Robust} \\ \midrule
Power-SLIC & \textbf{\checkmark}      & \textbf{\checkmark}   & \textbf{\textbf{\checkmark} }                & \textbf{\textbf{\checkmark} }                     \\
Varane     & \textbf{\checkmark}      & -   & \textbf{\textbf{\checkmark} }                & -                               \\
ECCPD      & \textbf{\checkmark}      & -   & \textbf{\textbf{\checkmark} }                & \textbf{\textbf{\checkmark} }                     \\
ERS        & -      & \textbf{\checkmark}   & -                          & -                               \\
SNIC       & -      & \textbf{\checkmark}   & \textbf{\textbf{\checkmark} }                & -                               \\
GMMSP      & -      & \textbf{\checkmark}   & -                          & -                               \\
SLIC       & -      & \textbf{\checkmark}   & -                          & -                               \\
MSLIC      & -      & \textbf{\checkmark}   & -                          & -                               \\
NR-SLIC    & -      & \textbf{\checkmark}   & -                          & \textbf{\textbf{\checkmark} }                     \\
SLIC0      & -      & \textbf{\checkmark}   & \textbf{\textbf{\checkmark} }                & \textbf{\textbf{\checkmark} }                     \\
KIPPI      & -      & \textbf{\checkmark}   & -                          & -                               \\
SEEDS      & -    & \textbf{\checkmark}   & -                          & -                               \\ \bottomrule
\end{tabular}
}
\caption{Qualitative comparison of multiple superpixel algorithms.}
\label{tb:qualitative_comparison}
\end{table}\vspace*{-2ex}

\begin{figure*}
\begin{center}
 \includegraphics[width=1\textwidth,trim={0 0.5cm 0 0.5cm},clip]{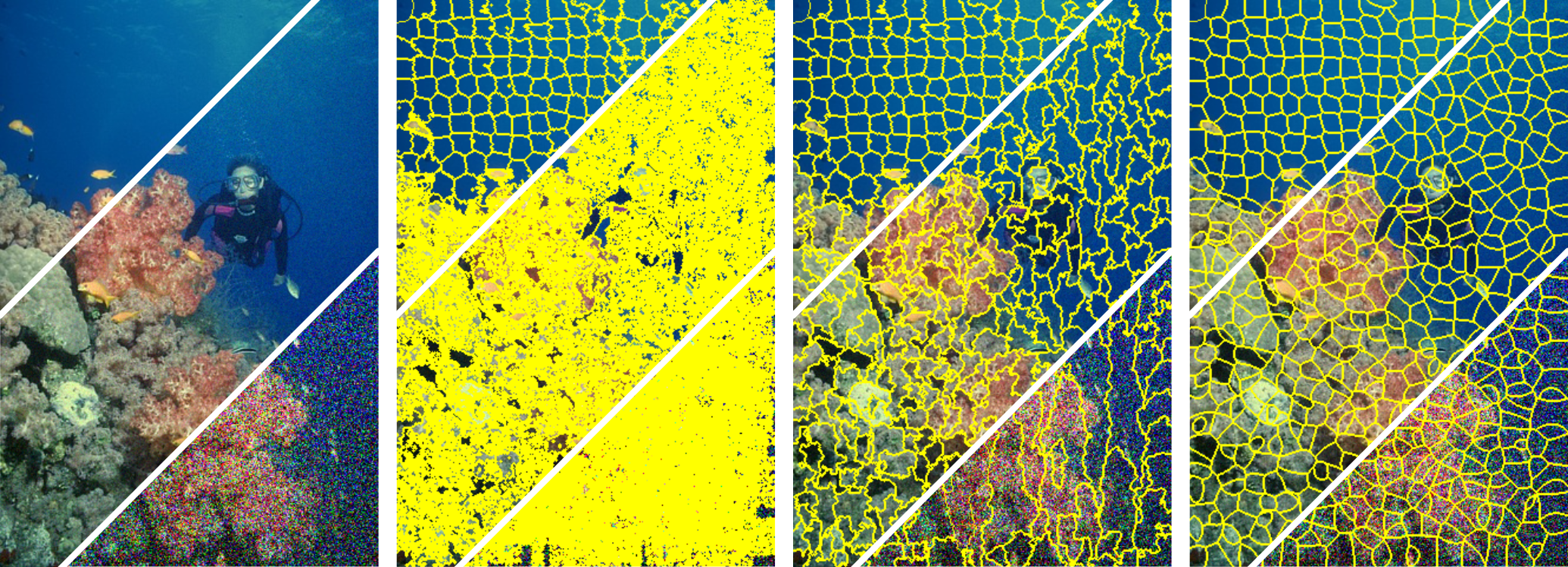}
\end{center}
\vspace{-0.1cm}
   \caption{Algorithms applied to an image with three levels of noise $\sigma^2 \in \{0,0.001,0.1\}$ and $k=700$. Left to right: original image, SLIC before post-processing, SLIC, and Power-SLIC.}
    \label{fig:compactness_small}
\end{figure*}

\paragraph{Qualitative Evaluation}


\Cref{tb:qualitative_comparison} provides a qualitative comparison of multiple superpixel algorithms. (Algorithms are classified as `diagram-based,' `fast,' `noise-robust,' and `compact' if their superpixels are described by diagram cells, their average running times for $k=500$ are below~1\si{s},  all their plots in \cref{fig:algorithm_comparison_and_noise} are close to horizontal lines, and their plots are in the top of the two groups for compactness in \cref{fig:algorithm_comparison_and_noise}.) \Cref{fig:algorithm_visual_comparison} gives a qualitative comparison of ten superpixel algorithms for varying levels of noise.

A comparison of the post-processing steps of Power-SLIC and SLIC is given in \cref{fig:compactness_small}. Although Power-SLIC uses SLIC's clustering approach, it is notable that the superpixels are fundamentally different due to the post-processing step.  This difference can be seen by comparing the results with the second column of \cref{fig:compactness_small}, which depicts the clustering result of SLIC without any post-processing. Only the local post-processing by SLIC shown in the third column creates a reasonable segmentation, which deteriorates for more extensive noise levels. In contrast, Power-SLIC optimizes more globally and returns a diagram representation. At the same time, this approach acts as a regularizer for high noise levels.

\section{Conclusion}
We presented a new superpixel algorithm, called Power-SLIC, whose superpixels are described by cells of diagrams. Using GBPDs, the quadratic boundaries appear to provide enough flexibility to capture complex image boundaries, while providing sufficient regularization to obtain very compact superpixels. This results in a high level of robustness under varying levels of additive white Gaussian noise. Additionally, Power-SLIC matches the running times of SLIC in theory and practice. Power-SLIC's added benefit of being resolution-independent can be used to speed up superpixel generation for larger images, as a full segmentation can be computed on a low-resolution version of the image.

{\small
\bibliographystyle{ieee_fullname}
\bibliography{biblio}
}

\end{document}